\documentclass{article}
\usepackage{spconf,amsmath,graphicx}
\usepackage{url}
\usepackage{subcaption}
\usepackage{multirow}
\usepackage{flushend}

\title{Unsupervised Domain Adaptation Schemes for Building ASR in Low-resource Languages}
\name{Anoop C S$^1$, Prathosh A P$^2$, A G Ramakrishnan$^1$}
\address{$^1$Indian Institute of Science, Bengaluru, India\\
  $^2$Indian Institute of Technology, Delhi, India}
\begin{document}
% \ninept
%
\maketitle
\begin{abstract}
Building an automatic speech recognition (ASR) system from scratch requires a large amount of annotated speech data, which is difficult to collect in many languages. However, there are cases where the low-resource language shares a common acoustic space with a high-resource language having enough annotated data to build an ASR. In such cases, we show that the domain-independent acoustic models learned from the high-resource language through unsupervised domain adaptation (UDA) schemes can enhance the performance of the ASR in the low-resource language. We use the specific example of Hindi in the source domain and Sanskrit in the target domain. We explore two architectures: i) domain adversarial training using gradient reversal layer (GRL) and ii) domain separation networks (DSN). The GRL and DSN architectures give absolute improvements of 6.71\% and 7.32\%, respectively, in word error rate over the baseline deep neural network model when trained on just 5.5 hours of data in the target domain. We also show that choosing a proper language (Telugu) in the source domain can bring further improvement. The results suggest that UDA schemes can be helpful in the development of ASR systems for low-resource languages, mitigating the hassle of collecting large amounts of annotated speech data 
\footnote{\label{myfootnote}Copyright 2021 IEEE. Published in the 2021 IEEE Automatic Speech Recognition and Understanding Workshop (ASRU) (ASRU 2021), scheduled for 14-18 December 2021 in Cartagena, Colombia. Personal use of this material is permitted. However, permission to reprint/republish this material for advertising or promotional purposes or for creating new collective works for resale or redistribution to servers or lists, or to reuse any copyrighted component of this work in other works, must be obtained from the IEEE. Contact: Manager, Copyrights and Permissions / IEEE Service Center / 445 Hoes Lane / P.O. Box 1331 / Piscataway, NJ 08855-1331, USA. Telephone: + Intl. 908-562-3966.}.
\end{abstract}
\section{Introduction}
\label{sec:intro}

Advancements in deep learning have brought performance improvements in acoustic and language modeling, yielding robust automatic speech recognition (ASR) systems in many languages. However, such systems require a large amount of speech data and the associated transcriptions. It is tough to collect large volumes of paired speech and transcriptions for most low-resource languages. It is estimated that only about 1\% of the world languages have the minimum amount of data that is needed to train an ASR \cite{low-resource}. However, in many cases, especially for languages in south Asia, we can find a ``close enough'' language with the same set (or a superset) of phonemes as the low-resource language and enough resources for building an ASR. In this work, we show that a better performing ASR can be built for the low-resource language using unsupervised domain adaptation (UDA) of acoustic models from the corresponding high-resource language. This method has the benefit of modeling on real data in comparison to the data augmentation techniques like vocal tract length perturbation \cite{vltp, vltp2}, speech and tempo perturbation \cite{audio_aug}, noise addition \cite{deepspeech}, data synthesis \cite{low_res_data_aug}, and spectral augmentation \cite{specaug}, where the modeling makes use of the synthetic data as well.

Unsupervised domain adaptation (UDA) has been successfully applied to various tasks to alleviate the shift between the train and test distributions. \cite{ganin} shows good adaptation performance in the classification task on digit image datasets having considerable domain shifts. They learn features that are discriminative for the image classification task and invariant to the domain. They introduce a gradient reversal layer (GRL) for achieving this objective. \cite{gender_accent_shift} reports performance improvements in speech recognition for data shifted in the domain by gender and accent. \cite{emo1} employs GRL layers to reduce the mismatch between train and test domains in the task of emotion recognition from speech data. \cite{speech} uses the GRL approach to improve the word error rate (WER) in speech recognition with the source domain data as clean speech and the target domain data as contaminated speech. They also show the robustness of the approach to the domain shifts caused by the differences in datasets.

The basic UDA scheme with a GRL tries to learn domain invariant features but ignores the individual characteristics of each domain. \cite{dsn} introduces domain separation networks (DSN) and shows improvements in a range of UDA scenarios in the image classification task. They learn two representations: one specific to each domain and the other common to both domains. \cite{dsn_clean2noisy} uses DSN for adaptation from clean speech to noisy speech. 

In this work, we explore the feasibility of the UDA schemes in the ASR task on low-resource languages that share the same acoustic space with a high-resource language. Specifically, we place the following assumptions in the selection of high and low resource language pairs. 
\begin{enumerate}
    \item The acoustic space spanned by the low-resource language is a subspace of that of the high-resource language. This requires the phoneme set of the low-resource language to be a subset of the high-resource language.
    \item There exists a high-resource language with reasonable amount of paired audio data and transcriptions to build an ASR.
    \item The low-resource language has enough text data available to train the language models.
    \item The speech data available in the low-resource language is quite limited, and the transcriptions are not available.
\end{enumerate}
Though the above assumptions seem quite constrictive, we can easily find a few low-resource languages in the Indian subcontinent which share a common acoustic space with a reasonably high-resource language. In this work, we use Hindi as the high-resource language and Sanskrit as the low-resource language, both belonging to the Indo-Aryan language family. Hindi is written in the Devanagari script, and many of its words are derived from Sanskrit. However, there exist substantial differences in the vocabulary and pronunciation between the two languages. One of the important differences is that the schwa, implicit in each consonant of the script, is not pronounced at the end of words and in some other contexts in Hindi. The script does not tell us when the schwa should be deleted, a phenomenon known as schwa deletion. For example, the word for "salty" is pronounced as  \textit{nam'kīn}  in Hindi and not \textit{namakīna}. Another difference is the pitch accents that are common in Sanskrit. Despite the above differences, these languages share a common phoneme set. So we can intuitively argue that there exists a domain shift between the distributions of acoustic features in Hindi and Sanskrit. This makes us believe that the speech recognition problem in Sanskrit, an extremely low-resource language, may be posed as an UDA problem from Hindi, a language with a reasonably good collection of annotated audio.

\section{Unsupervised domain adaptation for acoustic modeling}
\label{sec:uda}
We pose the problem of acoustic modeling in Sanskrit as an unsupervised domain adaptation task from Hindi. We build a deep neural network (DNN) - hidden Markov model (HMM) ASR system \cite{dnn-hmm} for Sanskrit with the domain-independent acoustic models learned from Hindi through UDA approaches. We make use of the UDA schemes introduced in \cite{ganin} and \cite{dsn}.
\begin{figure}[t]
  \centering
  \includegraphics[width=1\linewidth, height =0.6\columnwidth]{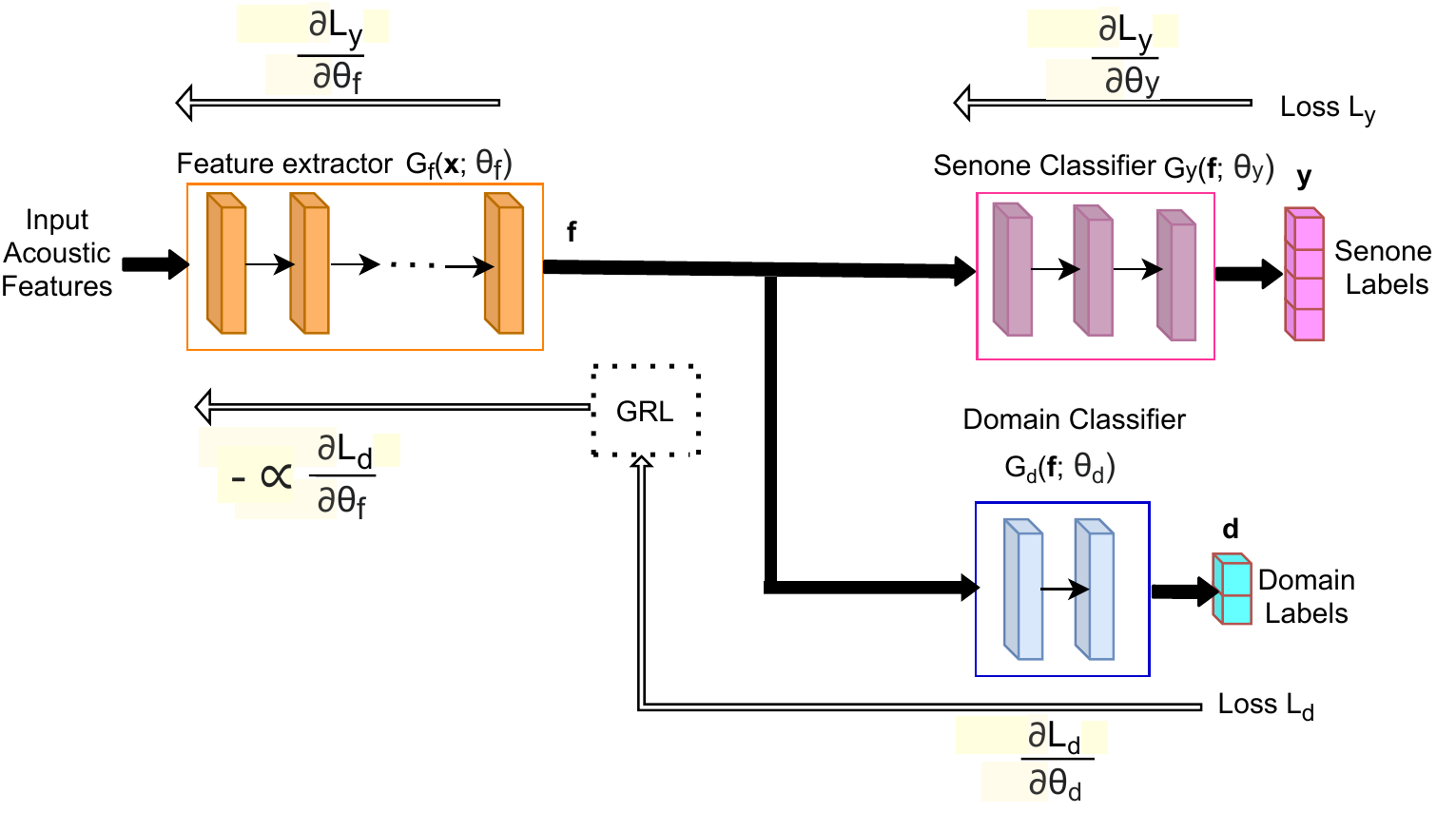}
  \caption{Block diagram of the basic scheme for unsupervised domain adaptation with a gradient reversal layer (GRL). $\theta_f$, $\theta_y$ and $\theta_d$ represent the parameters of the feature extractor, senone classifier, and domain classifier, respectively.}
  \label{fig:blockdiagram}
\end{figure}
\begin{figure*}[htb]
  \centering
  \includegraphics[scale=0.8]{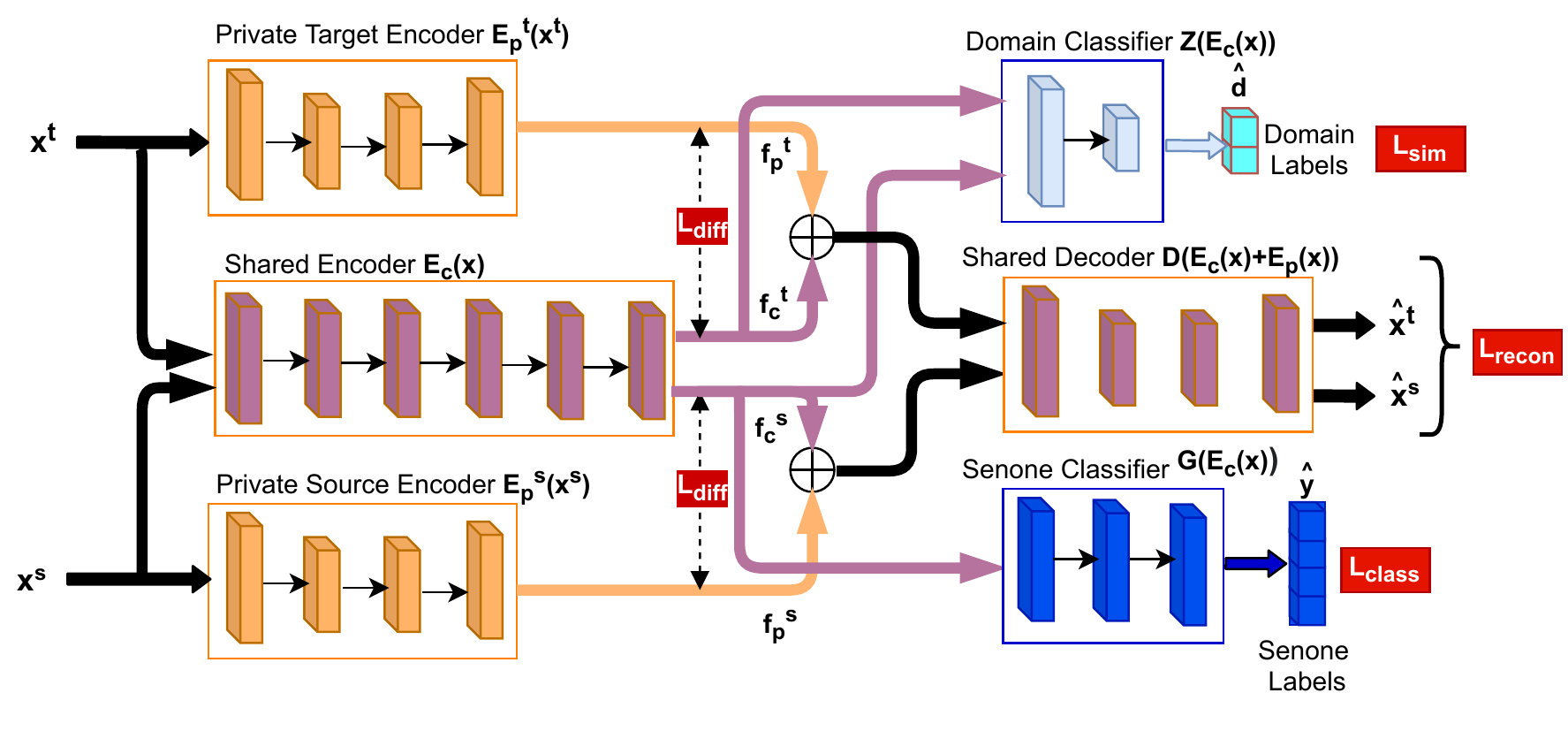}
  \caption{Block diagram of the domain separation network to model private and shared components. }
  \label{fig:dsn}
\end{figure*}
\subsection{Adversarial training using GRL}
\label{sec:grl}
In GRL-based adversarial training, we try to learn a feature representation invariant to the domain, but good enough in discriminating the senone labels. The neural network architecture for learning the acoustic model consists of three parts: feature extractor, senone classifier, and domain classifier. A block diagram of the UDA architecture employing GRL is shown in Fig. \ref{fig:blockdiagram}. Feature extractor $G_f$ maps the input acoustic features $\mathbf{x}$ to an internal representation $\mathbf{f} \in \mathcal{R}^D$. Senone classifier $G_y$ maps the output of the feature extractor to the senone labels $\mathbf{y}$ whereas the domain classifier $G_d$ maps them to domain labels $\mathbf{d}$. 
% \begin{eqnarray}
%     \mathbf{f} = G_f(\mathbf{x};\theta_f)\\
%     \mathbf{y} = G_y(\mathbf{f};\theta_y)\\
%     \mathbf{d} = G_d(\mathbf{f};\theta_d)
% \end{eqnarray}
% where $\theta_f$, $\theta_y$ and $\theta_d$ represent the parameters of the feature extractor, senone classifier, and domain classifier, respectively. 

We train the network to minimize the senone classification loss during the training phase by optimizing the parameters in the feature extractor $G_f$ and senone classifier $G_y$. This makes the network look for features that are capable of discriminating the senone labels. To make the features domain-invariant, we optimize the parameters of the feature extractor to maximize the domain classification loss. But at the same time, the parameters of the domain classifier $G_d$ are optimized to minimize the domain classification loss.  
This is achieved by introducing a GRL between the feature extractor $G_f$ and domain classifier $G_d$. During the forward pass, GRL acts as an identity transform. GRL takes the gradient from the subsequent layer during the backward pass, multiplies it with $-\alpha$, a hyperparameter to control the trade-off between senone discrimination and domain-invariance, and passes it to the preceding layer.  
% With the GRL the partial derivatives of $L_d$ with respect to $\theta_f$ gets multiplied by $-\alpha$ % which helps in implementing \eqref{eqn:f_update} with SGD. 

During the inference time, the domain classifier and GRL are ignored. The acoustic feature vectors are passed through the feature extractor and the senone classifier, and senone labels are predicted.
\subsection{Domain separation networks}
\label{sec:dsn}

A block diagram of the DSN architecture is shown in Fig. \ref{fig:dsn}. They model both the private and shared components of the domain representation. Private encoders $E_p^t(.)$ and $E_p^s(.)$ extract components $f_p^t$ and $f_p^s$ which are specific to the target and source domains. Shared encoder $E_c(.)$ is common to both domains and extract the shared components $f_c^t$ and $f_c^s$ . Shared decoder $D$ tries to reconstruct the input using the private and shared components. $G(.)$, the senone classifier maps $f_c^s$ to the senone label $\mathbf{\hat{y}}$. The domain classifier $Z(.)$ maps the shared components $f_c^s$ and $f_c^t$ to their respective domain labels $\mathbf{\hat{d}}$.

The network is trained to minimize the following loss function with respect to the parameters of $E_p$, $E_c$, $G$ and $Z$:
\begin{equation}
    L = L_{class} + {\beta} L_{sim} + {\gamma} L_{diff} + {\delta} L_{recon}
\end{equation}
where $\beta, \gamma$, and $\delta$ are hyperparameters. $L_{class}$ represents the senone classification loss and is applied only to the source domain. It is computed as the negative log-likelihood of the ground-truth senone labels. $L_{sim}$ represents the domain adversarial similarity loss and is computed as the negative log-likelihood of the domain labels. $L_{sim}$ ensures that the shared components $f_c^t$ and $f_c^s$ are as similar as possible irrespective of their domain so that the domain classifier cannot reliably predict the domain of the sample from its shared representation. Parameters of the domain classifier $Z(.)$ are trained to minimize the domain classification loss while the parameters of the shared encoder $E_c$ are trained to maximise the domain classification loss. This is also accomplished with a GRL. $L_{diff}$ encourages the shared component $f_c$ and the private component $f_p$ to encode different aspects of the input. This is achieved by imposing a soft subspace orthogonality constraint between the private and shared components.
\begin{equation}
    L_{diff} =  \| \mathbf{F}_c^{sT} \mathbf{F}_p^s \|^2_F + \| \mathbf{F}_c^{tT} \mathbf{F}_p^t \|^2_F 
\end{equation}
where $\mathbf{F}_c^{s}$, $\mathbf{F}_c^{t}$, $\mathbf{F}_p^s$ and  $\mathbf{F}_p^t$ are matrices with $f_c^s$, $f_c^t$, $f_p^s$ and $f_p^t$ as rows.  $\|.\|_F$ denotes the Frobenius norm. $L_{recon}$ is the reconstruction loss, computed as the mean squared error (MSE) between $\mathbf{x}$ and $\hat{\mathbf{x}} = D (E_c(\mathbf{x})+E_p(\mathbf{x}))$.
\begin{equation}
    L_{recon} = \sum_{i=1}^{N_s} \| \mathbf{x}_i^s - \hat{\mathbf{x}}_i^s\|^2 + \sum_{i=1}^{N_t} \| \mathbf{x}_i^t - \hat{\mathbf{x}}_i^t\|^2
\end{equation}
where $N_s$ and $N_t$ represent the number of speech frames from the source and target domains. We also validate the performance of the system with scale-invariant mean squared error (SIMSE) which is computed as:
\begin{equation}
    L_{SIMSE} = \frac{1}{k} \| \mathbf{x} - \hat{\mathbf{x}} \|_2^2 - \frac{1}{k^2} (\left[\mathbf{x} - \hat{\mathbf{x}}\right] . \mathbf{1}_k)^2
\end{equation}
where $k$ in the dimension of the input vector $\mathbf{x}$, $\mathbf{1}_k$ is a $k$-dimensional vector of ones and $\|.\|_2$ is the $L_2$ norm.

\section{Experimental setup}
\label{sec:setup}
\subsection{Datasets used for the study}
We primarily use a Hindi dataset \cite{data_hindi} in the source domain. We also use a Telugu dataset \cite{data_telugu} in the ablation studies. Both these datasets are the same as the ones used in multilingual and code-switching ASR challenge, Interspeech-2021. The details of both these datasets are available in \cite{multilingual}. Both the speech data and the corresponding transcriptions are available. 
% The dataset consists of train and test sets with 95.05 and 5.55 hours of audio, respectively. 
Hindi and Telugu audio files have sampling frequencies of 8 and 16 kHz, respectively. Both have 16-bit encoding. Telugu audio is downsampled to 8 kHz in our experiments. We randomly select 15,000 utterances ($\approx$ 15 hours) from their train sets for training the domain-independent acoustic models. The senone labels required for training the acoustic model are obtained from the alignments generated by a HMM-GMM system \cite{hmm} trained using Kaldi \cite{kaldi}. We also use a random selection of 1000 utterances from the test set to validate the domain independence of learned features. We refer to this set as \textit{dev}. 

The Sanskrit dataset used in the target domain has 3395 utterances with 16 kHz sampling frequency and 16-bit encoding. The data is randomly divided into two sets - train and test, with approximately 5.5 hours (2837 utterances) in the train set and 1 hour (558 utterances) in the test set. The train set is used for domain training, and the test set is used for inference. The data is downsampled to 8 kHz before its use in all our experiments. The text corpus for building the Sanskrit language models makes use of the \textit{wiki} Sanskrit data dump \cite{Sanskrit} and data from several Sanskrit websites. The extracted text is cleaned to remove unwanted characters and pre-processed to restrict the graphemes to the Devanagari Unicode symbols. 

\subsection{Details of feature extraction}
We use 40-dimensional filterbank features together with their delta and acceleration coefficients. Cepstral mean variance normalisation is performed at the utterance level. The features are spliced with a left and right context of 5 frames each. Thus the acoustic feature vector at the input of the DNN has dimensions of 1320 (40 x 3 x 11). Feature extraction is performed using Kaldi \cite{kaldi}.

\subsection{Training of the GRL model}
The feature extractor ($G_f$) has six hidden layers with 1024 nodes in each layer. The input to the $G_f$ is a 1320-dimensional acoustic feature vector, and the output is a 1024-dimensional feature $\mathbf{f}$. The feature vector $\mathbf{f}$ is forwarded to both the senone classifier $G_y$ and the domain classifier $G_d$. The senone classifier has two hidden layers, each with 1024 nodes and an output layer with 3080 nodes (equal to the number of senones in the Hindi training data). Domain classifier has a hidden layer with 256 nodes and an output layer with two nodes corresponding to the source and target domains. All the hidden layers are followed by batch normalization and ReLU activation. The logarithm of the softmax is computed at the output of both domain and senone classifiers. All the parameters ($\theta_f$, $\theta_y$ and $\theta_d$) are updated during the training with Hindi utterances. Only $\theta_f$ and $\theta_d$ are updated during the training with unlabelled Sanskrit utterances.

Negative log-likelihood loss is used for training. The models are trained using stochastic gradient descent with momentum \cite{sgd}. We use a batch size of 32 and an initial learning rate of 0.01. The learning rate is scaled by a factor of 0.95 after every 20000 steps. Training is performed for 20 epochs. The same number of frames from the source and target domains are used for training at every epoch. The domain adaptation factor $\alpha$ is gradually changed from 0 to 1 using the approach in \cite{ganin}.
% \begin{equation}
%     \label{eqn:alpha}
%     \alpha_{p} = \frac{2}{1+exp(-10*p)}-1
% \end{equation}
% Here $p$ varies from 0 to 1 over the epochs.

\subsection{Training of the DSN model}
Acoustic frames from the source and target domains, both of dimension 1320, are Input to the DSN. Private encoders for the source and target have four hidden layers with 512 nodes in each layer. The shared encoder has six hidden layers with 1024 nodes each. The senone and domain classifiers have the same architecture as the GRL model. All the hidden layers are followed by batch normalization and ReLU activation. The shared decoder has three hidden layers and an output layer with 1320 nodes.

The  hyperparameters $\beta$, $\gamma$, and $\delta$ are chosen as 0.25, 0.075, and 0.1, respectively. In order to promote the learning of the senone classifier in the initial phase of training, domain adversarial similarity losses are activated only after 10000 steps. The rest of the training process is the same as in the GRL model. 
% The domain adaptation factor $\alpha$ is modified as in \cite{ganin}. %\eqref{eqn:alpha}.

%\subsection{Training procedure}

\subsection{Decoding}
During the inference stage, only the output of the senone classifier is considered. The pre-softmax output of the senone classifier is normalized using the log probability of priors. % as in \eqref{eqn:prior_norm}. 
To find the most probable word sequence, we use weighted finite-state transducers (WFST) \cite{wfst} based decoding. 

The vocabulary of Sanskrit is distinct from that of Hindi. So the FSTs for grammar (G) and lexicon (L) are built using the text corpus collected in Sanskrit (target domain). Pronunciation dictionary for building the L-FST uses the grapheme to phoneme (G2P) mapping scheme in Sanskrit. This is different from the Hindi G2P scheme in aspects like schwa deletion and pronunciation of visargas \cite{asr_sanskrit}. HMM (H) and context-dependency (C)  FSTs are created using the HMMs learned from the source domain data. These four FSTs are composed to form a single HCLG graph, which maps the senones directly to the words in the target domain.

\section{Results}
\label{sec:results}
We decode the Sanskrit test set of 558 utterances with the adversarially trained GRL and DSN architectures.  These models use both the labeled Hindi data and the unlabelled Sanskrit data for training. In order to benchmark the performance of these UDA models, we decode the utterances using a simple DNN model trained only with the Hindi speech data. In this model, the domain classifier is not part of the network architecture. We also compare our results with a DNN model trained in multi-task (MT) learning setup, training the whole network to minimize both the senone and domain classification objectives. This network has the same architecture as the network in Fig. \ref{fig:blockdiagram}, except that it does not have the GRL. This model also makes use of both the Hindi and Sanskrit data. 

\begin{table}[htb]
\caption{WER on the Sanskrit-test set. Column-I gives the WER when the text corpus  for creating L and G FSTs includes the \textit{wiki} data dump and web-crawled data ($\approx$ 436840 words). Column-II gives the WER when the text corpus is restricted to the transcriptions of the Sanskrit speech corpus ($\approx$ 12250 words). }
\begin{center}
  \begin{tabular}{ | c | c | c | c | c |}
     \hline
    \textbf{Model}  & \textbf{Source} & \textbf{Target} & \textbf{I} & \textbf{II}\\ \hline
     DNN & Hindi & - & 24.58\% & 16.14\% \\ \hline
     MT & Hindi & Sanskrit & 21.43\% & 13.10\% \\ \hline
     GRL & Hindi & Sanskrit & 17.87\% & 10.22\% \\ \hline
     DSN & Hindi & Sanskrit & 17.26\% & 9.89\% \\ \hline
     \hline
  \end{tabular}
\end{center}
\vspace{-0.2cm}
\label{tab:wer}
\end{table}

The performance measure used to evaluate these models is word error rate (WER). The results are shown in Table \ref{tab:wer} for two cases: i) when the language models are trained on the whole text corpus made out of \textit{wiki} data dump and web-crawling (column I) and ii) when the language models are trained only on the transcriptions of the Sanskrit speech corpus (column II).

GRL approach gives an absolute improvement of 6.71\% in WER over the baseline DNN model when the language model is trained on a large text corpus in Sanskrit (column I of Table \ref{tab:wer}). DSN provides an absolute improvement of 7.32\%. Both the UDA models have nearly similar performance, and they beat the multi-task learning model by more than 3.5\%. When the language model training is restricted to the transcriptions of the Sanskrit speech corpus, the performance of all the models improves as expected. However, the UDA approaches still retain the edge over the MT models by about 3\%. We have also tried to fine-tune the UDA models using the senone labels of the Sanskrit train set computed from the DNN models trained on Hindi, but they did not improve the performance of the models.

\subsection{Ablation studies}
In all the experiments below, we compute WER using the bi-gram language models trained on the larger text corpus associated with column I of Table \ref{tab:wer}.

\subsubsection{Effect of the amount of unlabelled training data from the target domain}
\begin{figure}[htb]
  \centering
  \includegraphics[width=1\linewidth]{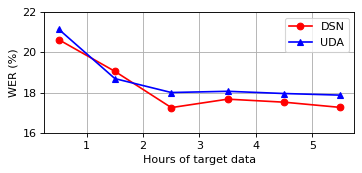}
  \caption{Effect of the amount of unlabelled training data from the target domain on the WER. }
  \label{fig:hours}
\end{figure}

Next, we address the question of the amount of unlabelled data required for proper adaptation. We split the training set in the target domain into six sets with 0.5, 1.5, 2.5, 3.5, 4.5, and 5.5 hours of data and train the UDA models using them. The entire source domain data is used for training. The performance of these models in terms of WER is shown in Fig. \ref{fig:hours}. The performance of both the models improves as the amount of unlabelled training data increases but nearly saturates after about 2.5 hours of data in the target domain (Sanskrit). 
\vspace{-0.7cm}
\subsubsection{Domain independence of features}
\begin{figure}[htb]
\begin{minipage}[b]{0.48\linewidth}
\centering
\includegraphics[width=\textwidth]{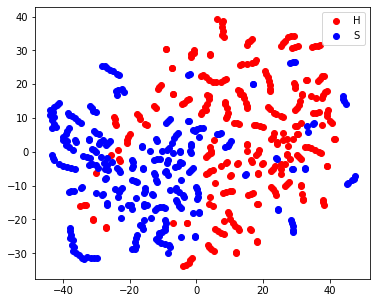}
\vspace{-0.6cm}
\label{fig:tsne_dnn}
\subcaption[]{}
\end{minipage}
\hspace{0.1cm}
\begin{minipage}[b]{0.48\linewidth}
\centering
\includegraphics[width=\textwidth]{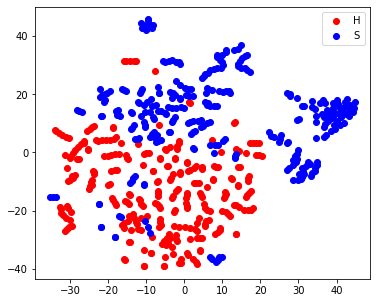}
\vspace{-0.6cm}
\label{fig:tsne_mt}
\subcaption[]{}
\end{minipage}
% \vspace{-1cm}
\begin{minipage}[b]{0.48\linewidth}
\centering
\includegraphics[width=\textwidth]{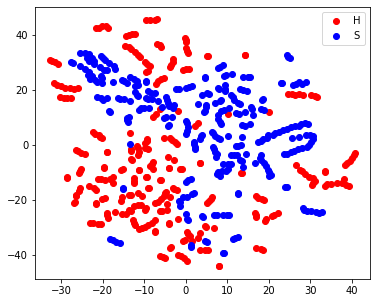}
\vspace{-0.6cm}
\label{fig:tsne_grl}
\subcaption[]{}
\end{minipage}
\hspace{0.1cm}
\begin{minipage}[b]{0.48\linewidth}
\centering
\includegraphics[width=\textwidth]{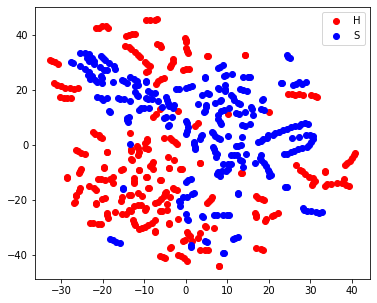}
\vspace{-0.6cm}
\label{fig:tsne_dsn}
\subcaption[]{}
\end{minipage}
\caption{2-D t-SNE plots of features at the output of feature extractor/shared encoder in (a) baseline DNN (trained only with Hindi data), (b) MT, (c) GRL, and (d) DSN models for frames with senone label 3009. \textit{H}  and \textit{S} denote the frames obtained from the Hindi-dev set (source domain) and Sanskrit-test set (target domain), respectively.}
\label{fig:tsne}
\end{figure}

We also visualize the features at the output of the feature extractor (shared encoder for DSN) for our models. We collect an equal number of frames with the same senone label (based on the HMM-GMM alignment) from the Hindi-dev and Sanskrit-test sets and plot the vectors at the output of the feature extractor (or shared encoder) using t-SNE \cite{tsne}. The results are shown in Fig. \ref{fig:tsne}. Compared to Figs.\ref{fig:tsne}(a) and \ref{fig:tsne}(b), the features in Figs. \ref{fig:tsne}(c) and \ref{fig:tsne}(d) are more intermingled indicating the domain independence of features. The domain-discriminative power is higher for the features from MT models as seen from Fig. \ref{fig:tsne}(b).

To verify the extent of domain independence achieved by the model, we compute the frame-level domain accuracy on the Hindi-dev and the Sanskrit-test sets. The results are listed in Table \ref{tab:domain_acc}. It can be seen that the features from UDA models are much more domain-independent compared to the MT models.

\begin{table}[htb]
\caption{Frame-level domain accuracy of the UDA models computed on the Sanskrit-test and Hindi-dev sets.}
\label{tab:domain_acc}
\centering
\begin{tabular}{|c|c|c|}
\hline
      & \multicolumn{2}{c|}{\textbf{Domain Accuracy}} \\ \hline
\textbf{Model} & \textbf{Sanskrit-test}           & \textbf{Hindi-dev}           \\ \hline
MT    & 91.86\%                 & 76.31\%             \\ \hline
GRL   & 63.77\%                 & 44.94\%             \\ \hline
DSN   & 63.21\%                 & 52.18\%             \\ \hline
\hline
\end{tabular}
\end{table}
\vspace{-0.6 cm}
\subsubsection{Effect of the loss functions in DSN}
Next, we experiment with the constituents of the loss function in DSN. We train four models: (i) with all the loss functions in DSN, (ii) without the difference loss ($\gamma$ = 0), (iii) without the similarity loss ($\beta$ = 0), and (iv) with the reconstruction loss computed as SIMSE. The results are listed in Table \ref{tab:dsnloss}. The performance degrades slightly in the absence of difference loss which tries to enhance the orthogonality between the private and shared components. There is considerable degradation in the performance in the absence of similarity loss. The results are still better than the baseline DNN and MT models (refer column I in Table \ref{tab:wer}) indicating the usefulness of private and shared component decomposition in DSN. The model using SIMSE as the reconstruction loss performs inferior to that of the one using MSE. 
\begin{table}[htb]
\caption{Effect of the different constituents of the loss function on the performance of the DSN.}
\begin{center}
  \begin{tabular}{ | l | c | }
     \hline
    \textbf{Loss functions}  & \textbf{WER} \\ \hline
     All terms included &  17.26\% \\ \hline
     $L_{diff}$ = 0 ($\gamma$ = 0) & 18.10\% \\ \hline
     $L_{sim}$ = 0 ($\beta$ = 0) & 20.37\% \\ \hline
     $L_{recon}$ = SIMSE & 18.00\% \\ \hline
     \hline
  \end{tabular}
\end{center}
\vspace{-0.5cm}
\label{tab:dsnloss}
\end{table}

% We have further explored the trade-off between domain-invariance and the performance of the DSN in terms of WER. We compute the frame-level domain accuracy and WER for various values of $\beta$ and the results are shown in Figure.
\vspace{-0.6cm}
\subsubsection{Source domain language selection}
Though both Hindi and Sanskrit are written in Devanagari, they have a difference in pronunciation, as pointed out in section \ref{sec:intro}. However, \cite{telugu} suggests that Telugu and Malayalam are the closest languages to Sanskrit in terms of pronunciation, vocabulary, and grammar.  Moreover, the schwa is retained in Dravidian languages like Telugu and Malayalam, just like Sanskrit. Since suitable datasets are available in Telugu, we repeat the experiment with Telugu in the source domain, hoping for better acoustic and pronunciation models in Sanskrit. The results are shown in Table \ref{tab:telugu}. All the models improve, as the phone HMM models learned from Telugu better match Sanskrit. Here also, the performance of UDA models is better than the DNN and MT models. The UDA approaches improve by around 3.5-4.5\% compared to adaptation from Hindi.

\begin{table}[htb]
\caption{WER on the Sanskrit-test set when trained with Telugu as the source domain language. }
\centering
\begin{tabular}{|c|c|c|c|}
\hline
\textbf{Model} & \textbf{Source} & \textbf{Target}   & \textbf{WER}     \\ \hline
DNN   & Telugu & -        & 17.65\% \\ \hline
MT    & Telugu & Sanskrit & 14.71\% \\ \hline
GRL   & Telugu & Sanskrit & 13.09\% \\ \hline
DSN   & Telugu & Sanskrit & 13.72\% \\ \hline
\hline
\end{tabular}
\label{tab:telugu}
\end{table}
\vspace{-0.6cm}
\section{Conclusions}
\label{sec:conclusions}
In this work, we propose UDA as an option to tackle the scarcity of data in low-resource languages which share a common acoustic space with a high-resource language. We experiment with Hindi as the source domain language and Sanskrit as the target domain language. GRL and DSN models improve the WER by 6.71\% and 7.32\%, respectively, compared to a baseline DNN model trained only on Hindi. The models perform better than the multi-task learning framework. Proper selection of source domain language (Telugu in our case) further improves the results. The results indicate that UDA can provide a faster way of building ASR systems in low-resource languages, reducing the hassle of collecting large amounts of annotated training data if a suitable high-resource language is available.
\bibliographystyle{IEEEbib}
\bibliography{refs}
\end{document}